\pdfoutput=1

\documentclass[11pt]{article}

\usepackage[final]{acl}

\usepackage{times}
\usepackage{soul}
\usepackage{url}
\usepackage[utf8]{inputenc}
\usepackage{graphicx}
\usepackage{amsmath}
\usepackage{amsthm}
\usepackage{enumitem}
\usepackage{booktabs}
\usepackage{algorithm}
\usepackage{algorithmic}
\usepackage[switch]{lineno}
\usepackage{cleveref}
\usepackage{multirow}
\usepackage{latexsym}
\usepackage{color}
\usepackage{pifont}
\usepackage{xcolor}
\usepackage[normalem]{ulem}
\usepackage{tabu}
\usepackage{times}
\usepackage{latexsym}
\usepackage{tabularx}
\usepackage[T1]{fontenc}
\usepackage[utf8]{inputenc}

\usepackage{microtype}
\usepackage{inconsolata}

\title{Guided Profile Generation Improves Personalization with LLMs}

\author{Jiarui Zhang \\
  University of Southern California \\
  \texttt{jzhang37@usc.edu} \\}

\begin{document}
\maketitle
\newcommand{\tocite}{{\color{red}CITE} }
\newcommand{\toref}{{\color{red}REF} }
\newcommand{\jiarui}[1]{{\color{red}(JR: #1)}}

\newcommand{\pc}{\texttt{PC}}
\newcommand{\pp}{\texttt{PP}}
\newcommand{\gpg}{\texttt{GPG}}

\newcommand{\cmark}{\ding{51}}%
\newcommand{\xmark}{\ding{55}}%

\begin{abstract}

In modern commercial systems, including Recommendation, Ranking, and E-Commerce platforms, there is a trend towards improving customer experiences by incorporating Personalization context as input into Large Language Models (LLMs). However, LLMs often struggle to effectively parse and utilize sparse and complex personal context without additional processing or contextual enrichment, underscoring the need for more sophisticated context understanding mechanisms. In this work, we propose Guided Profile Generation (\gpg{}), a general method designed to generate personal profiles in natural language. As is observed, intermediate guided profile generation enables LLMs to summarize, and extract the important, distinctive features from the personal context into concise, descriptive sentences, precisely tailoring their generation more closely to an individual's unique habits and preferences. Our experimental results show that~\gpg{} improves LLM's personalization ability across different tasks, for example, it increases $37\%$ accuracy in predicting personal preference compared to directly feeding the LLMs with raw personal context. 

\end{abstract}


\begin{figure*}[t]
    \centering
    \includegraphics[trim=0 0 0 0, clip, width=0.95\textwidth]{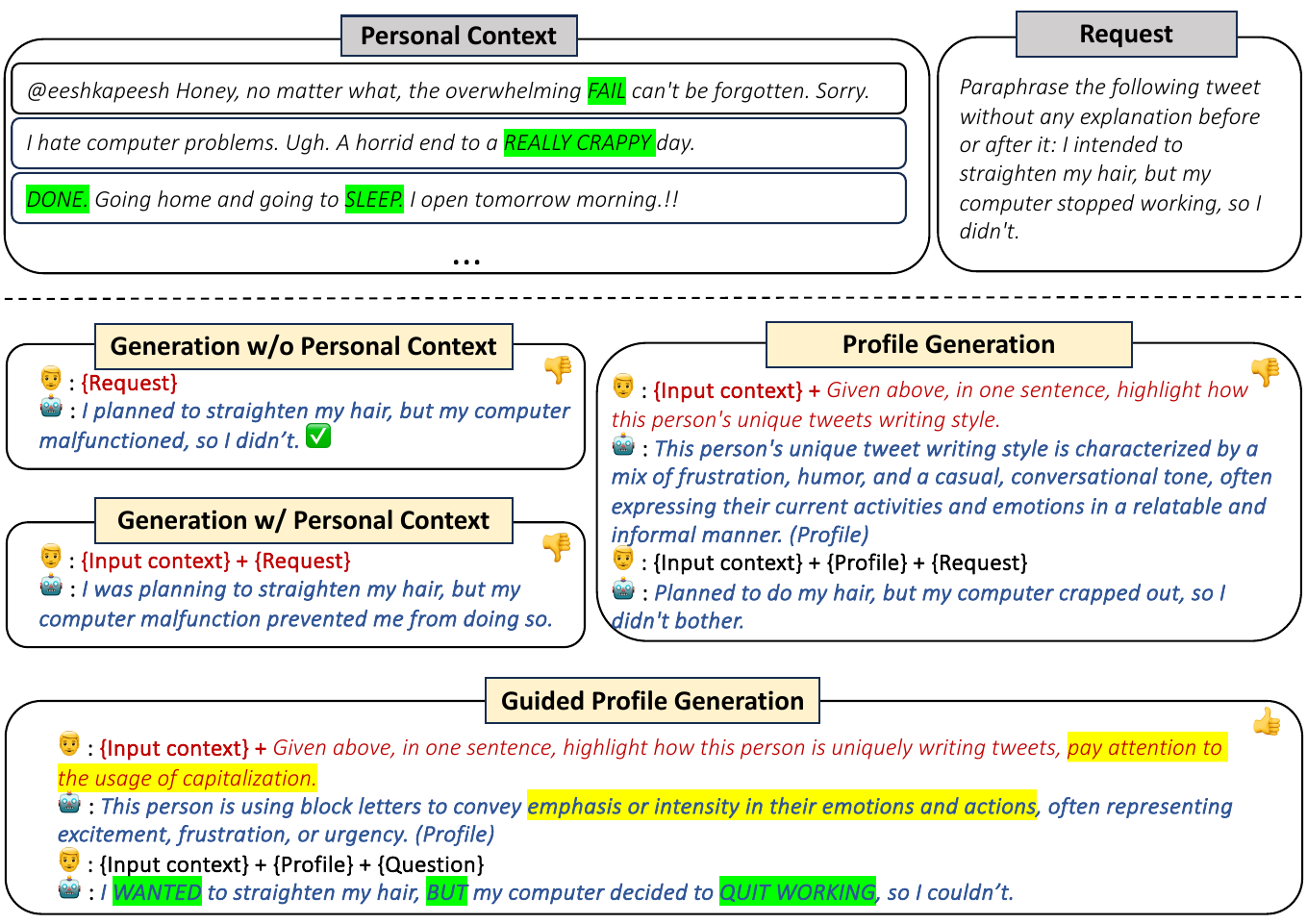}
    \caption{A motivating example. The model is given a personal context reflecting the person's writing style, and the task is to paraphrase a new tweet for the user. We show~\texttt{gpt-3.5-turbo-1106}'s response under different input conditions. The result shows that generating a descriptive personal profile with proper guidance helps the model finish the personalization better.}
    \label{figure:motivation}
\end{figure*}
\section{Introduction}
\label{sec:intro}

Within the context of personalization tasks, personal profiling has been extensively employed. Conventional methodologies typically rely on substantial datasets such as graph-based similarities. These profiles often exhibit `neighborhoods' and `relationships' within the data, posing challenges for immediate interpretability without supplementary processing.
Recently, LLMs have demonstrated robust capabilities in tasks related to reasoning and generation, leading to a growing interest in leveraging LLMs for personalization services. 
However, distinguished from other Naturual Language Processing (NLP) tasks, we identify two primary challenges in personalization with LLMs.

The first challenge is the complexity of personal contexts and the sparsity of their key information. For example, a person's distinctive writing style may only be discernible in a small portion of their writing, whereas the remainder of the writing style tends to be more generic. As is shown in recent studies~\cite{lost_in_the_middle}, LLMs have challenges in capturing comprehensive information within lengthy contexts, making it easy to overlook the smaller portions that contain distinctive writing styles. Previous studies~\cite{rag,lamp} have attempted to address this challenge by context retrieval. However, context retrievers frequently rely on surface-level ranking strategies, such as keyword similarity. Such an approach, while straightforward, may not always align with the nuanced needs of personalization tasks.

The second challenge lies in the balance between generalization and personalization. While LLMs have demonstrated considerable performance on general tasks, they still struggle to generate output that fully aligns with users' desired behaviors and directions~\cite{llm_struggle1}.
Rather, they prioritize imitating the majority of their training sets ~\cite{gpt_not_succeed}. 
 Figure ~\ref{figure:motivation} illustrates a personalized task involving the paraphrasing of a tweet to match someone's distinctive writing style. From the personal context, it is noticeable that the individual tends to use block letters to emphasize actions and feelings. However, the model closely mirrors the original question input when receiving the personal context and question directly, which can be reachable even without personal context. When we instruct LLM to describe the person's writing style, rather than noticing the spatial use of capitalization, it pays attention to the emotion, and content, which are not our desired `writing styles'. 

Steering LLM outputs precisely is always a challenge. To address it, previous work has attempted to apply reinforcement learning from human feedback~(RLHF)~\cite{instructgpt}. However, this is a resource-intensive process that might be financially burdensome and impractical for some service providers. Other works tried to train compact models~\cite{stimulousprompting} from the feedback of comparison between LLM's output and ground truth labels. However, no certain true label is available for independent profile generation tasks. 
Prompt optimization, involving both manual and automated efforts in designing and selecting suitable prompts for various tasks, stands out as a promising and widely adopted alternative.

The majority of recent studies on prompt optimization indicate that LLMs can benefit from digesting intermediate generated prompts to successfully complete complex tasks~\cite{cot,zscot,react}. In personalization, formulating a personal profile serves as a crucial intermediate step that enhances task performance in terms of both accuracy and efficiency.
Most existing profile modeling techniques depend on substantial datasets. While these approaches are effective for structured analysis, they often yield profiles that require additional interpretation. Additionally, these profiles tend to be restricted to a limited range of data types, limiting the inclusion of more diverse perspectives. In contrast, natural language is not only inherently understandable and easily diagnosable, but it also enables the expansion of the scope of data types that can be effectively integrated into the modeling process.

In this paper, we propose a general method leveraging LLMs for personalization, named~\textbf{Guided Profile Generation~(\gpg{})}, whose goal is to augment LLMs' capacity for interpreting raw personal contexts and to generate high-quality natural language personal profiles. In~\gpg{}, the process begins with personal context digestion, where we pose specific questions in predetermined directions on personal context. Then the model will generate descriptive natural language personal profiles, steered by the output of last stage. The resulting personal profile will be subsequently employed to respond to the request with downstream models.

We conduct extensive experiments to evaluate the efficacy of~\gpg{} with ~\texttt{gpt-3.5} on the task of purchase preference prediction, text paraphrasing, and dialogue response generation and benchmark the performance of~\gpg{} with several baselines. Our result shows that~\gpg{} consistently enhances the personalization performance across various tasks. 
In preference prediction of online purchase,~\gpg{} improve $37\%$ accuracy in predicting personal preference of product purchasing compared with direct prediction with raw context.
In text paraphrasing on Tweet,~\gpg{} improves METEOR score by $2.24$ by digesting the writing style with the recognition of the most significant writing features. 
Furthermore, we conduct ablation studies to evaluate the impact of various components within the~\gpg{} framework and undertake further analysis to comprehend the limitations of our methods, aiming to pave the way for future directions in this research.

\section{Related Work}

LLMs have demonstrated robust performance through scaling up, in-context learning~\cite{gpt3}, reinforcement learning from human feedback~\cite{instructgpt}, and instruction tuning~\cite{instructtune}, making them capable of complex reasoning tasks~\cite{mmlu,bigbench,brainteaser1,brainteaser2}. The performance of the model is sensitive to input and output manners, making prompt optimization~\cite{react,cot,zscot,planningediting} a popular topic.

There has been a growing interest in using LLMs for personalization. LLM-Rec~\cite{llmrec}  utilizes LLMs as recommenders by prompting them with recommendation instructions and employing graph-based engagements. However, this approach lacks emphasis on the crafting of user profiles. LAMP~\cite{lamp} attempts to integrate a context retriever to avoid the need for feeding the entire personal context to LLMs, but the retrieved personal context still proves challenging for LLMs to easily comprehend. PALR~\cite{palr} uses LLMs to generate user profiles for personalized recommendation and fine-tuned llama~\cite{llama} to generate ranking. However, the exploration of more effective methods for crafting user profiles in natural language based on personal contexts with diverse structures remains underexplored. Other studies also explore the use of LLMs to augment graph-based recommendation system~\cite{llmrec}, support human writing creativity~\cite{Creativitysupport}, personalized writing education~\cite{writingassistant2}, dialogue systems~\cite{dialogue1} and healthcare assistant~\cite{hearlthcare}. 

For datasets, LAMP~\cite{lamp} introduces seven language tasks that necessitate personalization. These tasks include tweet paraphrasing and email subject generation, among others. Notably, tweet paraphrasing serves as a comprehensive test bed for evaluating personalized writing style imitation using LLMs. Amazon review~\cite{amazonreview} provides abundant online purchase history and shopping reviews, enabling the creation of a preference prediction dataset for product purchasing. PER-CHAT~\cite{dialoguedata} is an open-domain single-turn dialogue dataset collected from Reddit. In PER-CHAT, each dialogue response is paired with related comment history from the same user, enabling personal profile crafting. Other datasets like MovieLens~\cite{movielens}, Recipe~\cite{receipts}, PERSONA-CHAT~\cite{PERSONA-CHAT} are also widely used. We evaluate~\gpg{} by personalized preference prediction, tweet paraphrasing, and dialogue generation sets in this paper.

\section{Guided Profile Generation}
\label{sec:gpg}

\begin{figure*}[t]
    \centering
    \includegraphics[trim=0 0 0 0, clip, width=0.95\textwidth]{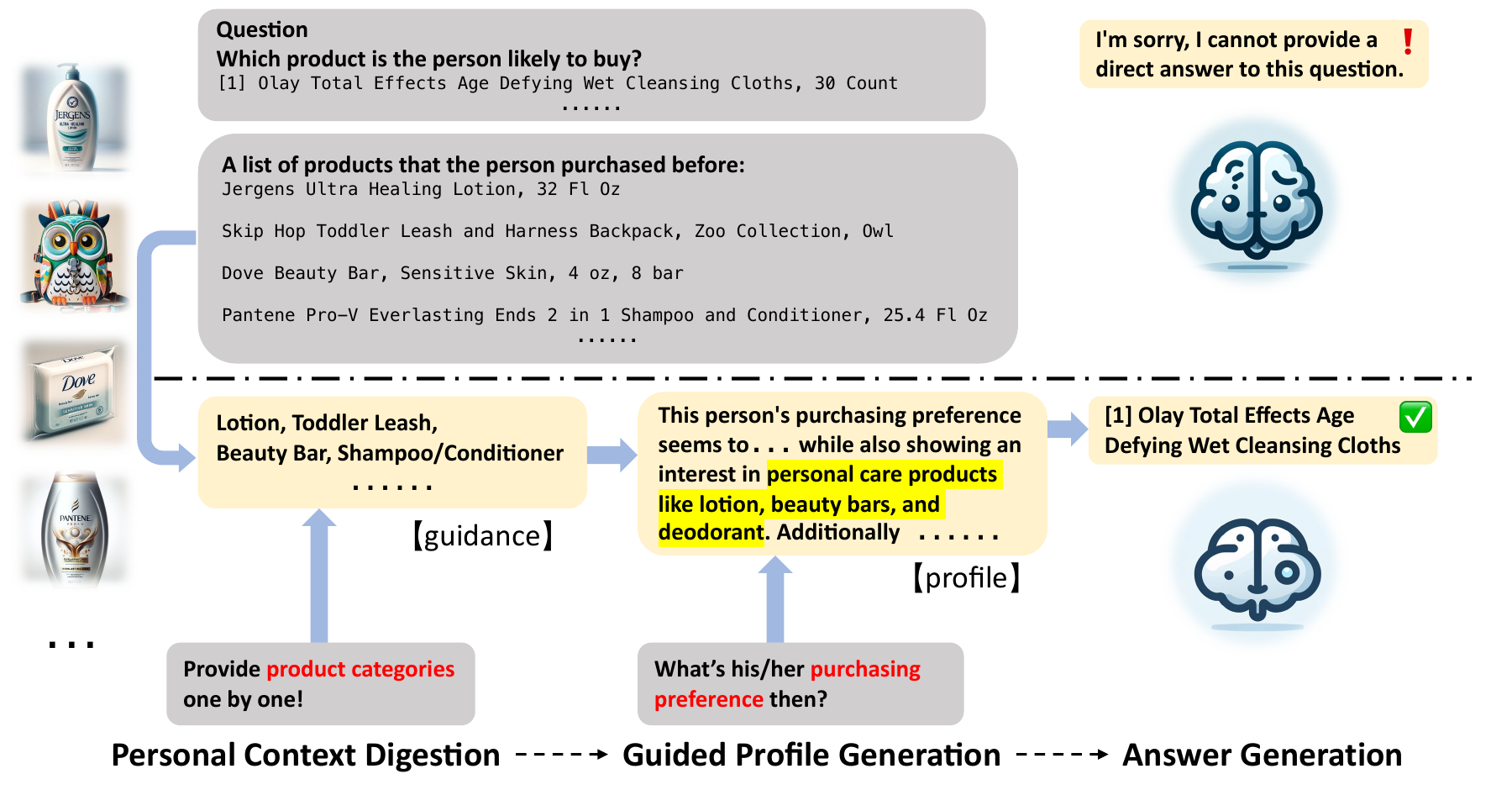}
    \caption{Illustration of~\gpg{}  described in Section~\ref{sec:gpg}: Given a personal context, we instruct LLM to generate a descriptive personal profile via self-guidance. The personal profile is then used to complete the personal task.~\gpg{} enables LLM to generate high-quality personal profiles, improving their performance on personalization. Note that our experiments are conducted in~\textbf{textual domain}, images are for illustrative purposes.}
    \label{fig:pipeline}
\end{figure*}
Given a personal context~\pc{}, and a task~\texttt{T}, the objective of personalization is to align with the individual's behavior and successfully accomplish the task. In contemporary commercial systems, personal profile crafting proves advantages for both accuracy and efficiency, achieved by providing a clear reflection of a person's behavior and ensuring reusability without the need to process the raw context again. Given the impressive capabilities of LLMs, there is a natural inclination to leverage them for integrating raw~\pc{} and generating personal profiles. However, our early investigation indicates that these approaches may not achieve the expected performance (\autoref{figure:motivation}). Moreover, the lack of human-annotated data for intermediate personal profiles makes direct optimization through fine-tuning a challenging option.

We propose~\gpg{}, a general method for personalization with LLMs through personal profile generation. The proposed method of~\gpg{} is presented in~\autoref{fig:pipeline}. Different from joint learning with downstream personalization tasks for LLM, which adopts Reinforcement Learning from Human Feedback (RLHF), we adopt a much more cost-effective yet efficient method. This method focuses on generating a readable, descriptive personal profile based on personal context.
Our method consists of the following steps:

\subsection{High-level Workflow}
The first step is Personal Context Digestion. In this step, we pose task-specific questions to the LLM, guiding it to digest~\pc{} in our desired direction. For instance, in the scenario of predicting a customer's preferred product based on their purchase history, we prompt the model to sequentially generate product categories. 
The main purpose of this step is to get direction and key information for the next step.
Note that differentiated from few-shot prompting which needs a large amount of in-context corpus crafted by humans, in~\gpg{}, only one specific question is designed for each task.

The second step is Guided Profile Generation. The response of the previous steps serves as guidance for the generation of the personal profile. Similar to~\cite{stimulousprompting}, 

we concatenate the~\pc{} and guidance as input. We instruct the LLM to generate descriptive sentences serving as the personal profile. In contrast to high-dimensional representations, our profile is explainable, enabling easy diagnosis of inadequacies. Moreover, our profile is language model orthogonal, facilitating broader applications and seamless future development.

The final step is Response Generation. The generated personal profile is used to finish the final task. To provide sufficient information, we do not exclude the raw personal context in our main experiment. In~\cref{sec:discussion}, we conduct a detailed experiment study of the effect of the inclusion of~\pc{} and guidance.

\section{Evaluation Tasks and Metrics}
Our proposed method can be applied to a wide range of personalization tasks to overcome the challenge given by raw personal contexts. In this work, we mainly focus on the task of personalized preference prediction, text paraphrasing, and dialogue continuation.

\subsection{Task of Preference Prediction}

In commercial systems, accurately predicting a user's preference is one of the most crucial tasks. This prediction holds the potential to benefit various downstream tasks (e.g., personal recommendation). However, reliance on large databases and specific models, like assessing the similarity between different users, poses limitations. The design of these models often restricts access to additional information, such as the full name and detailed product information on the Internet.
Furthermore, these large databases are not always readily accessible for common use. In contrast, LLMs exhibit the capability to process any textual data, providing a means to overcome the aforementioned limitation. In this section, we delve into their ability to predict user preferences relying solely on textual data. Specifically, we choose user-based online purchase history as our focus due to the distinctive personal behaviors evident in this domain.

Specifically, to construct the test bed for user preference prediction, we leverage the Amazon Product Review~\cite{amazonreview,amazonreview2} dataset collected from the Amazon website. The dataset provides the purchase history for each of product with categories and users. We extract the purchase history for each of user and keep the product categories. Then we filter out all of the users who have purchased less than 5 categories of product, who are considered as being lack of personal context. For the remaining users, we randomly select one of the purchased product categories with at least 2 products. Then one of the products is selected as a question. To sample the distractors, we randomly select 3 products from the category that this person has never purchased before. We consider the product name to be enough information to identify the person's purchase preference, to the end, we exclude all of the review information in the dataset for simplicity. 
In the resulting dataset,~\pc{} is defined as purchasing history, which is a list of products that the person has purchased before, and the task is to identify the product that is most likely to be purchased by the person, and select the product from four candidate options.

\textbf{Metrics.} Since the dataset is in the form of multiple choice questions, and is designed to be in a balanced set, we take the accuracy as the only metric for this task. Lastly, it is worth noting that the constructed preference prediction dataset mostly serves as a diagnosis purpose, evaluating how we can better utilize LLMs predicting user's preference based on raw context. As is shown in~\autoref{tab:amazon}, a single semantic-level comparison algorithm can reach the highest performance in such data, but will not generalize well when facing different formats of datasets.

\subsection{Task of Text Paraphrasing}
\label{sec:textpara}
Though simple for humans, it is underexplored whether LLMs can detect and imitate the text-writing styles for different individuals. Such capability is crucial since in recent times, LLMs have been widely used as writing assistants. In this section, we explore how well can LLMs imitate a person's writing style given the raw~\pc{}. Compared to formal writing, such as news reports or research articles, Twitter is a platform where every individual can express their thoughts freely. Hence, we select the text on Twitter as our study focus due to the frequently personalized writing on it, such as punctuation, and abbreviations. Specifically, we use LAMP-7~\cite{lamp}, a user-based Twitter collection based on sentiment140~\cite{rawtwitter} dataset. In LAMP-7, one of a user's tweets is selected as the source of task input. Then, this input is fed into an LLM for neutralizing the writing style. 
In the resulting dataset,~\pc{} is defined as the collection of all past tweets that this person had before excluding the selected one. The task is to reconstruct the tweet following this person's writing style based on the neutralized tweet and all other tweets.

\textbf{Metrics.} We consider the word and phrase level usage similarity, including BLEU~\cite{bleu}, METEOR~\cite{meteor} and ROUGE~\cite{rouge}. Since the task is style reconstruction without semantic-level personalization, we do not evaluate the semantic-level (embedding) similarity.

\subsection{Task of Dialogue Response Generation}

Besides writing style imitation discussed in~\cref{sec:textpara}, the ability of AI assistants to accurately reflect an individual's opinion is also crucial. This task is particularly challenging due to the opinions are often implicit and multifaceted in a raw personal context, and should be selectively employed based on the requirements of different tasks.

We focus on dialogue continuation in practice. In particular, we leverage PER-CHAT~\cite{dialoguedata} collected from open-domain discussions on Reddit. PER-CHAT collects each individual's comment history, and the task is to use the history as a signal of personal preference and help the individual answer the question. We do not include the retrieved personal profile from the paper for simplicity. To improve the relevance between the comment history and target response, we measure their semantic similarities based on sentence-transformer~\cite{sentencetransformer}, and select a subset having a maximum similarity larger than 0.4. We also exclude instances with max similarities larger than 0.6 to avoid overlap between comment history and target response.

\textbf{Metrics.}
We consider semantic level similarity metric based on sentence-transformer and BERT-Score~\cite{sentencetransformer,bertscore} as our main metric for evaluation. Since the posted questions are mostly open-ended discussions without definite answers, we do not include metrics for direct string, word or phrase-level comparison.

\section{Experiments}
\label{sec:discussion}

We use OpenAI's~\texttt{gpt-3.5-turbo-1106} as our major LLM all through the tasks; during inference, we keep the temperature at 0 (greedy decoding) to gain a deterministic result and set max\_tokens to 100. We report the result with a single run due to the greedy decoding.

\begin{table}[t]
\small
\caption{Accuracy comparison of different prompting strategies on amazon preference prediction dataset. Where~\texttt{DG} denotes direct generation,~\texttt{PG} denotes profile generation directly with language instructions.}
\label{tab:amazon}
\centering
\begin{tabular}{lc}
\toprule
 \textbf{Method}& \textbf{Accuracy} \\ \hline
\texttt{Random} & 25.00 \\\hline
\texttt{DG w/o PC}& 31.65 \\
\texttt{DG w/ PC} & 47.55 \\
\texttt{PG} & 54.98 \\
\gpg & \bf 65.08\\
\bottomrule
\end{tabular}
\end{table}

\begin{figure*}[t]
    \centering
    \includegraphics[width=0.95\textwidth]{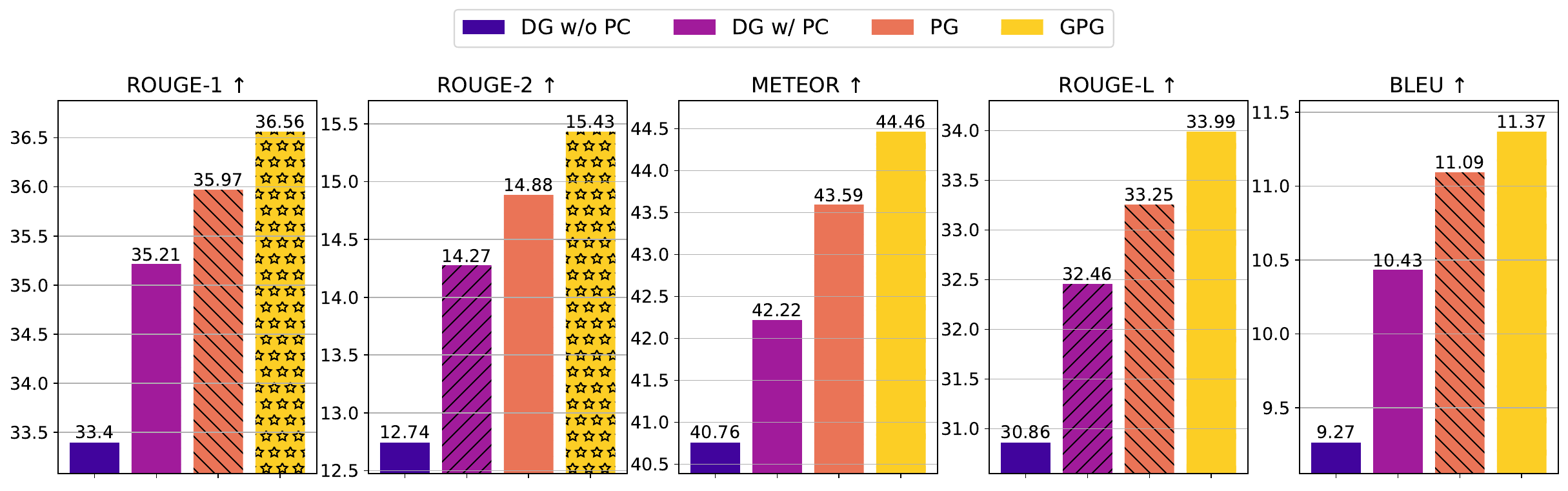}
    \caption{
    Text paraphrasing on Twitter performance of~\gpg{} in comparison with direct generation without personal context (\texttt{DG w/o PC}), direct generation with personal context (\texttt{DG w/ PC}) and Profile Generation (\texttt{PG}).}
    \label{fig:tweet}
\end{figure*}

\begin{table}[t]
\small
\caption{Performance of different prompting strategies on our selected subset of PER-CHAT data, where~\textbf{ST} denotes sentence transformer and~\textbf{BS} denotes Bert-Score.}
\label{tab:dialogue}
\centering
\begin{tabular}{lcc}
\toprule
\textbf{Method} & \textbf{ST}& \textbf{BS} \\ \hline
\texttt{DG w/o PC}& 29.86  & 83.09 \\
\texttt{DG w/ PC} & 32.31  & 83.54 \\
\texttt{PG}       & 32.66  & 83.47 \\
\gpg              & 32.35  & 83.43 \\
\bottomrule
\end{tabular}
\end{table}

\subsection{Baselines.}
For the comparison purpose, we present the following baselines to illustrate the effectiveness of~\gpg{}:
\begin{enumerate}[leftmargin=*,itemsep=1pt,topsep=0pt,parsep=0pt]
    \item \textbf{Direct Generation without Personal Context.}~(\texttt{DG w/o PC}) We consider the LLMs' native response to the question since they have been trained on numerous corpus. For example, LLMs could have knowledge about the general tweet writing style, thus having the ability to reshape a sentence to such a style. The input is formalized as~\texttt{\{Q\}}.
    \item \textbf{Direct Generation with Personal Context.}~(\texttt{DG w/ PC}) In this baseline, we feed the~\pc{} to LLM and ask them directly to generate the answer to our question. The input is formalized as~\texttt{\{PC\}\{Q\}}.
    \item \textbf{Unguided Profile Generation.}~(\texttt{PG}) In this baseline, we ask LLMs to generate the profile of a person according to~\pc{} without further instructions. Then we use the generated profile to finish the personalized task. The input is formalized as~\texttt{\{PC\}\{PP\}\{Q\}}, where~\texttt{PP} is the profile generated from~\pc{} by instructing LLM.
\end{enumerate}

\subsection{\gpg{} Specifications.}
\label{sec:specification}
In the task of Preference Prediction, we guide the LLM to generate the personal profile by providing the product categories. To this end, we first ask the LLM~\textit{``Provide the product category of above one by one, each of them use less than 10 words, split by a comma:''}. The resulting list of categories serves as the guidance for LLM in generating the personal profile. After the generation of the personal profile, we concatenate the raw~\pc{}, and the personal profile as the final input of LLM, predicting the final answer. We do not include the raw guidance, i.e. purchase category to reduce redundant information. We will discuss the effect of the inclusion of each component in detail in~\cref{sec:ablation}.

In the Text Paraphrasing task, the LLMs are guided by a unique aspect of the writing style of the tweets when generating the personal profiles. We identify 4 key aspects of paraphrasing:~\textit{Capitalization, Emoji, Abbreviation, Punctuation}. Then we instruct LLM to select the most distinctive features in the personal context, specifically our instruction is:~\textit{Among the usage of 1. Capitalization, 2. Emoji, 3. Abbreviation, 4. Punctuation, which is the most distinctive feature of the above tweets?}. Then LLM will generate the profile based on the self-selected category and use the generated profile together with the guidance to finish the task.

In the Dialogue Response Generation Task, We expect the generated personal profile to be a summary of these texting habits and personal opinions. Inspired by the original paper, we instruct LLM to generate the basic personal information from their comment history, the aspects include:~\textit{“pets”, “family”, “residence”, “favorites”, “partner”, “possessions”, “gender”, “self-description”}. Then the above aspects are used to craft the personal profile.

\subsection{Experimental Results}
Table~\ref{tab:amazon} shows the performance on our Amazon preference prediction dataset of different prompting strategies. LLM improves its performance by~$50.23\%$ when adding the personal context to its input. Furthermore, this improvement can be further enhanced to~$73.71\%$ by using a self-generated personal profile. Our~\gpg{}, reaching an improvement of~$105.62\%$ through self-guidance.

The result of tweet paraphrasing is shown in~\autoref{fig:tweet}. 
Firstly, the inclusion of personal context improves the performance across all metrics, clearly showing the usefulness of personal context in reshaping the users' writing styles. 
Generating an unguided personal profile further improves the performance compared to direct generation, providing guidance could double such benefit. Such a result indicates the effectiveness of generating a self-guided intermediate profile for personalization of text paraphrasing with LLMs.

On dialogue generation, the inclusion of raw~\pc{} has a positive impact on the performance, as is shown in~\autoref{tab:dialogue}. However, profile generation, either guided or unguided does not help much in such a task. To understand this phenomenon better, we will look deeper into the generations in~\cref{sec:ablation}.

\begin{table*}[]
\small
\caption{The ablation study on amazon preference prediction~(\textbf{P-P}) and text paraphrasing~(\textbf{T-P}) tasks. We consider the inclusion of raw personal context~(\pc{}), guidance~(\texttt{G}, context digestion), and descriptive personal profile~(\texttt{PP}). The best performances are in bold.}
\label{tab:ablation}
\centering
\begin{tabular}{cccc|c|ccccccc}
\toprule
&\multicolumn{3}{c}{Dataset} & \textbf{P-P} & \multicolumn{5}{c}{\textbf{T-P}} \\
&w/~\pc{}? & w/~\texttt{G}? & w/~\texttt{PP}? & \textbf{Acc} & \textbf{ROUGE-1} & \textbf{ROUGE-2} & \textbf{METEOR}& \textbf{ROUGE-L} &\textbf{BLEU} \\
\midrule
\multirow{2}{*}{\texttt{DG}} 
&\cmark & -  & - & 47.55 & 35.21 & 14.27 & 42.22 & 32.46 & 10.43  \\
&\xmark & -  & - & 31.65 & 33.40 & 12.74 & 40.76 & 30.86 & 9.27  \\
\midrule
\multirow{2}{*}{\texttt{PG}}
& \cmark & - & \cmark & 54.98 & 35.97 & 14.88 & 43.59 & 33.25 & 11.09 \\
& \xmark & - & \cmark & 51.86 & 34.25 & 13.57 & 42.04 & 31.65 & 9.95 \\
\midrule
\multirow{6}{*}{\gpg{}}
& \cmark & \xmark & \cmark & \textbf{65.08} & 36.12 & 15.14 & 43.87 & 33.55 & 11.23 \\
& \xmark & \xmark & \cmark & 58.25 & 33.96 & 13.43 & 43.50 & 31.41 & 10.10 \\
\cmidrule(lr){2-10}
& \cmark & \cmark & \cmark & \underline{61.96} & \textbf{36.56} & \textbf{15.43} & \textbf{44.46} & \textbf{33.99} & \textbf{11.37} \\
& \xmark & \cmark & \cmark & 59.14 & 35.90 & 14.62 & 44.45 & 33.32 & 10.81 \\
\cmidrule(lr){2-10}
& \cmark & \cmark & \xmark & 51.71 & 35.69 & 14.75 & 43.07 & 33.11 & 10.79 \\
& \xmark & \cmark & \xmark & 48.44 & 35.04 & 13.84 & 42.52 & 32.42 & 10.10 \\

\bottomrule
\end{tabular}

\end{table*}

\section{Analysis and Discussions}

\subsection{Ablation Studies}
\label{sec:ablation}

We conduct an ablation study to better understand the benefit of each component of~\gpg{}, on preference prediction and text paraphrasing tasks. the result is shown in~\autoref{tab:ablation}. 
Specifically, we analyze the impact of incorporating personal context~(\pc{}), guidance(\texttt{G}, context digestion), and personal profile~(\texttt{PP}) during the generation of \textbf{final response}. Next, we will provide a detailed analysis based on the result.

\noindent
\textbf{Can we exclude raw personal context when generating an answer?}
In our experiment, we initially incorporated the personal context as part of the input to mitigate the risk of information loss. However, in practice, it is inefficient to keep the personal context as input during every run. To this end, we remove the personal context during the final task generation.
Compared with the direct generation,~\gpg{} improve the performance by $17.53\%$ (absolute) in predicting purchase preference, generations without raw personal context (sixth-row in~\autoref{tab:ablation}) could approximate $61.04\%$ of such improvement, indicating a considerable trade-off between the expense and performance. However, in text paraphrasing, the performance after removing the raw personal context is worse than a direct generation, underlining the higher importance of personal context in text paraphrasing.

\noindent
\textbf{Can personal context digestion directly benefit the downstream tasks?}
As is shown by our result, personal context digestion can help LLMs generate better descriptive personal profiles. Thus, we are curious whether such a benefit is directly applicable to the final task generation. To this end, we skip the generation of descriptive personal profiles and directly perform downstream tasks after context digestion, the result is shown in the last two rows of~\autoref{tab:ablation}. Surprisingly, the guidance itself is functioning even worse than an unguided personal profile (third row) in both of the tasks, suggesting: \textbf{1.}~Despite being beneficial in enhancing the generation of personal profiles, the guidance itself is not immediately effective for improving the performance of the final task. \textbf{2.}~A descriptive personal profile helps the model be better at personalization. 

\subsection{Error analysis and Observations.}

\noindent
\textbf{Profile Generation helps LLM be more certain about making selections.}
We find LLMs frequently opt to abstain from responding when faced with uncertain information.
To better understand this behavior of LLMs, we select all of the `abstain' answers and report the ratios of correct, incorrect, and abstained answers in the preference prediction dataset. Specifically, the answer is recognized as abstained if the word `sorry' is found in the answer. From the result shown in~\autoref{tab:abstain}, we find that the primary improvement of~\gpg{} on preference prediction data is from helping the model reduce the ratio of answer abstaining rather than correcting their failures.

\begin{table}[t]
\small
\caption{The ratio of correct, incorrect, and abstain answers in the amazon preference prediction dataset.}
\label{tab:abstain}
\centering
\begin{tabular}{lccc}
\toprule
\textbf{Method} & \textbf{Correct} & \textbf{Incorrect} & \textbf{Abstain} \\ \midrule
\texttt{DG w/o PC} & 27.79 & 55.27 & 16.94 \\
\texttt{DG w/ PC} & 41.46 & 32.39 & 26.15 \\
\texttt{PG} & 52.30 & 34.92 & 12.78 \\
\texttt{\gpg{}} & 64.04 & 31.20 & 4.75 \\
\bottomrule
\end{tabular}
\end{table}

\subsection{Limitation and Future Works}

\noindent
\textbf{Integrating multiple aspects personalization.}
Our experiments are conducted on a single source of personal context.
In practice, the complete profile of an individual should be drawn from multiple aspects. For example, a person's purchase preference can be related to their gender, age, habit, or even the weather where they live. 
Due to the difficulty of cross-platform data collection, most of the off-the-shelf personalization data are from a single source. Constructing datasets containing personal contexts from multiple sources for each individual could be interesting. In addition, it is also challenging to integrate data from multiple aspects. While wisely designed mechanisms like graph contrastive learning~\cite{heterogeneous} could potentially incorporate different types of information, unifying graph information into natural language is a lightweight alternative~\cite{kgadapt,transferring}, obtaining better explainability at the same time. We believe our findings bring useful insight into this future direction.

\noindent
\textbf{Multimodal personalization.}
Recently, multimodal large language models~(MLLMs)~\cite{instructblip,llava} have shown promising capabilities in various tasks. Such advancement opens the possibility of multimodal personalization. For example, an individual's preference for clothes could be highly related to the designs, which are not easily described by text. In such studies, the undesired and generic MLLM outputs could be a problem, applying a visual crop~\cite{vicrop} directed by visual search~\cite{vstar} as a `guidance' would be interesting. In addition, other modalities such as sound~\cite{Seamless}, and sensor data like heart rates~\cite{heartrate} are also considerable.


\section{Conclusion}

In this work, we present Guided Profile Generation~\gpg{}, a novel method leveraging LLMs for personalization tasks through profile generation and context digestion. We conduct extensive experiments on various personalization tasks, including preference prediction, text paraphrasing, and dialogue continuation.
Despite the superior performance,~\gpg{} generates a personal profile in pure natural descriptive language, which is interpretable and easily diagnosable. Moreover, we reveal why and how the guidance and descriptive personal profile improve the performance. We hope our research can pave the way for personalization applications with AI models in the future.

\bibliography{ref}
\clearpage

\appendix

\section{Examples of Three Tasks}
\begin{figure*}[t]
    \centering
    \includegraphics[trim=0 0 0 0, clip, width=0.99\textwidth]{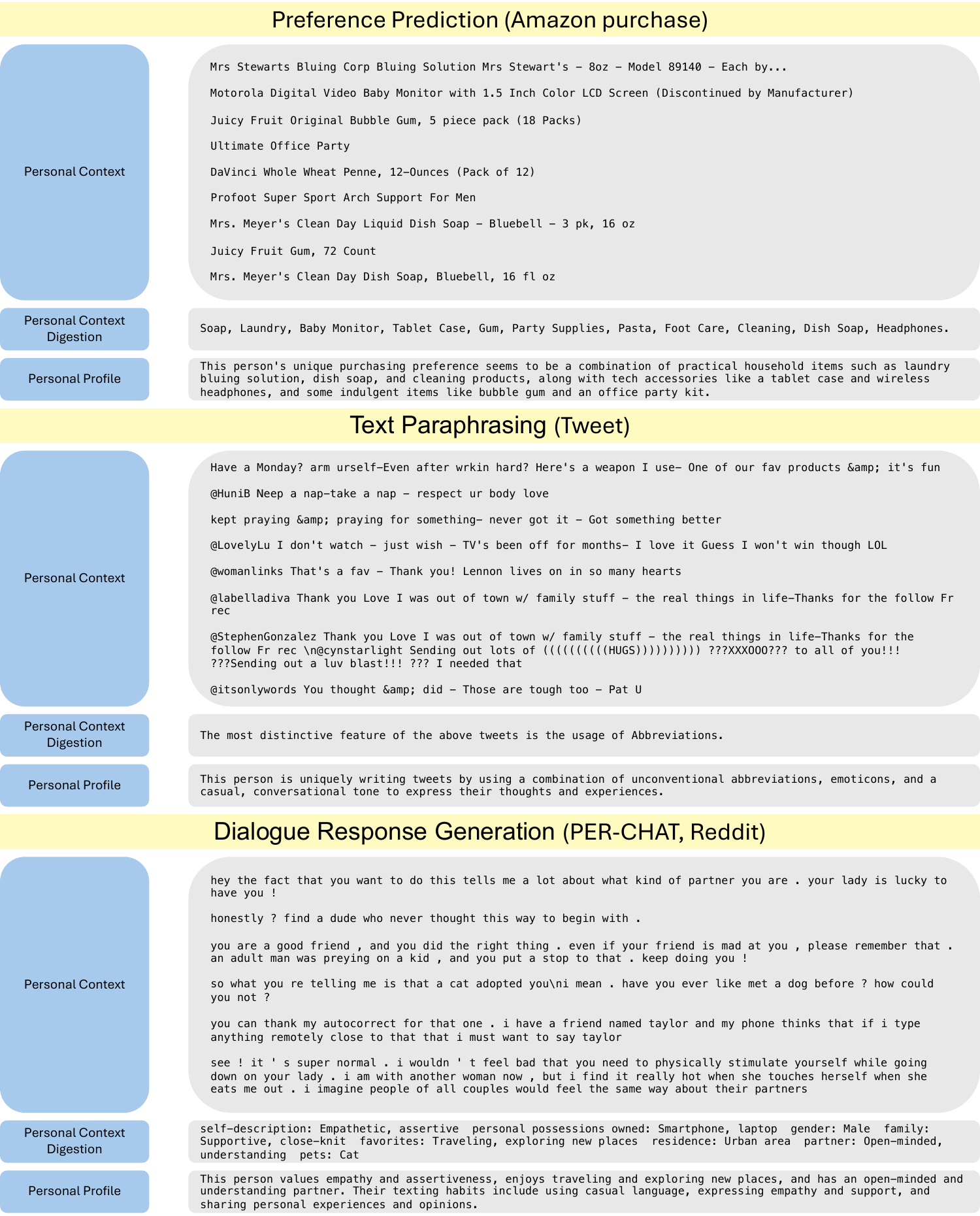}
    \caption{Examples of personal context, personal context digestion, and personal profile of three tasks under our test. We select only part of the personal context due to their length.}
    \label{figure:example}
\end{figure*}

In~\autoref{figure:example}, we present examples of the three tasks under our test, we include raw personal context, personal context digestion, and personal profile in each example. The prompts for generating personal context digestion and personal profiles can be found in~\cref{sec:specification}.

\section{Challenges in Open-Ended Personalization Tasks.}

\begin{table}
\centering
\small
\caption{One Example from dialogue continuation task, where LLMs face a challenge in closely aligning the personal preference and texting habits.}
\label{tab:polite}
\begin{tabular}{p{7.5cm}}
\tabucline[1.1pt]\\
\textit{Profile:} 
This person seems to have strong opinions and\\ enjoys giving advice. They also appear to be direct\\ and straightforward in their communication style,\\ often using humor and sarcasm. \\
\textit{Question (task):} 
How do you talk to a girl irl after only talking online? \\
\textit{Human (ground truth):} 
always through mutual acquaintances. Anything else is creepy and stalker like . \\
\textit{LLM response:} 
Well, first off, be yourself and don't try to be someone you're not. Keep the conversation light and fun, and don't be afraid to throw in some humor. And remember, confidence is key! Good luck! \\\tabucline[1.1pt]\\
\end{tabular}
\label{tab:error}
\end{table}
When addressing open-ended tasks like dialogue continuation, LLMs encounter more challenges in aligning with personal preferences and texting habits. One example is shown in~\autoref{tab:polite}, where LLMs are trying to give a generic response to the question rather than a personalized one. This tendency aligns with findings reported in~\cite{gpt_not_succeed} that LLM would prioritize imitating the majority of their training data. While such a phenomenon is not bad in itself as it helps LLMs leverage huge amounts of data and obtain impressive capabilities, it is not a desired behavior in personalization.

\section{Statistics of Three Tasks.}
~\autoref{tab:statistic} presents the statistics of three included tasks. We report the total count of data instances (\# Data) and the average number of user activities (\# Activities) within each personal context. Specifically, in the Preference Prediction task, \# Activities represents the average number of products a user has purchased before. In Text Paraphrasing, it represents the average number of history Tweets. In Dialogue Response Generation, it represents the average number of dialogue responses within the personal context.

\begin{table}[t]
\small
\caption{Statistics of preference prediction~(\textbf{P-P}), text paraphrasing~(\textbf{T-P}) and Dialogue Response Generation~(\textbf{D-G}). We report the total number of data (\# data) and the average number of user activities (\# Activities) per personal context.}
\label{tab:statistic}
\centering
\begin{tabular}{lccc}
\toprule
\textbf{Task} & \textbf{P-P} & \textbf{T-P} & \textbf{D-G} \\ \midrule
\texttt{\# Data} & 673 & 1500 & 607 \\
\texttt{\# Activities} & 6.82 & 17.64 & 10.00 \\
\bottomrule
\end{tabular}
\end{table}

\end{document}